\pdfoutput=1
\documentclass{article}

\usepackage{microtype}
\usepackage{graphicx}
\usepackage{hyperref}
\usepackage[accepted]{latex/icml2024}
\makeatletter 
\renewcommand{\ICML@appearing}{Pre-print.}
\makeatother

\usepackage{float,stfloats}
\usepackage{graphicx,tikz,relsize}
\usetikzlibrary{shapes.geometric, shapes.multipart, arrows, arrows.meta, positioning, calc, chains, backgrounds}
\usepackage{enumitem}
\usepackage[capitalize,noabbrev]{cleveref}
\usepackage[textsize=tiny]{todonotes}

\renewcommand{\cite}[1]{\textcolor{red}{WRONG CITE COMMAND [\StrSubstitute{\detokenize{#1}}{,}{, }]}}
\newcommand{\figurewidth}{\linewidth}

\icmltitlerunning{Interpretability Needs a New Paradigm}

\begin{document}

\twocolumn[
\icmltitle{Interpretability Needs a New Paradigm}

\begin{icmlauthorlist}
\icmlauthor{Andreas Madsen}{mila,poly}
\icmlauthor{Himabindu Lakkaraju}{harvard}
\icmlauthor{Siva Reddy}{mila,mcgill,cifar-fb}
\icmlauthor{Sarath Chandar}{mila,poly,cifar-ca}
\end{icmlauthorlist}

\icmlaffiliation{mila}{Mila}
\icmlaffiliation{poly}{Polytechnique Montréal}
\icmlaffiliation{harvard}{Harvard University}
\icmlaffiliation{mcgill}{McGill University}
\icmlaffiliation{cifar-fb}{Facebook CIFAR AI Chair}
\icmlaffiliation{cifar-ca}{Canada CIFAR AI Chair}

\icmlcorrespondingauthor{Andreas Madsen}{andreas.madsen@mila.quebec}

\icmlkeywords{Interpretability, Explanations, Transparency, Paradigms, Post-hoc, Intrinsic, Ethics, Future work, Faithfulness measurable models, Self-explanations, Self-explaining models}

\vskip 0.3in
]

\printAffiliationsAndNotice{}

\begin{abstract}
Interpretability is the study of explaining models in understandable terms to humans. At present, interpretability is divided into two paradigms: the intrinsic paradigm, which believes that only models designed to be explained can be explained, and the post-hoc paradigm, which believes that black-box models can be explained. At the core of this debate is how each paradigm ensures its explanations are \textit{faithful}, i.e., true to the model's behavior. This is important, as false but convincing explanations lead to unsupported confidence in artificial intelligence (AI), which can be dangerous. This paper's position is that we should think about new paradigms while staying vigilant regarding faithfulness. First, by examining the history of paradigms in science, we see that paradigms are constantly evolving. Then, by examining the current paradigms, we can understand their underlying beliefs, the value they bring, and their limitations. Finally, this paper presents 3 emerging paradigms for interpretability. The first paradigm designs models such that faithfulness can be easily measured. Another optimizes models such that explanations become faithful. The last paradigm proposes to develop models that produce both a prediction and an explanation.

\end{abstract}

\section{Introduction}

There was a time in physics, in the late 17th century, when Isaac Newton insisted that light is a particle and Christiaan Huygens insisted that light is a wave \citep{Huygens1690}. These ideas were seemingly irreconcilable at the time. Of course, now we have a much better theory, and we understand that light can be seen as both a wave and a particle.\footnote{Known as the wave-particle duality concept \citep{Messiah1966}.}

In 1874, Georg Cantor proposed set theory and showed there exists at least two kinds of infinity. This divided the mathematical field. The Intuitionists, who named Cantor’s theory nonsense, thought that math was a pure creation of the mind and that these infinities weren’t real. Henri Poincaré said: ``Later generations will regard Mengenlehre (set theory) as a disease from which one has recovered'' \citep{Poincare1993}. Leopold Kronecker called Cantor a ``scientific charlatan'' and ``corruptor of the youth'' \citep{Dauben1977}.

The other group, the Formalists, thought that by using Cantor’s set theory, all math could be proven from this fundamental foundation. David Hilbert said: ``No one shall expel us from the paradise that Candor has created'' and ``In opposition to the foolish Ignoramus (we will not know; i.e., intuitionists), our slogan shall be: We must know – we will know'' \citep{Smith2014}.

Today, we know infinities are important concepts; thus, the Intuitionists were wrong. However, Kurt Gödel showed that the Formalists were also wrong. Unfortunately, there exist true statements which can never be proven \citep[Gödel's incompleteness theorem]{Godel1931}.

These are just two examples in science and mathematics where there have been strong debates and beliefs due to conflicting paradigms. Science historian Thomas Kuhn defines a scientific paradigm as: ``universally recognized scientific achievements that, for a time, provide model problems and solutions to a community of practitioners'' \citep{Kuhn1996}.

Time and time again, when there are conflicting paradigms, it is only ``for a time''. Eventually, we find neither paradigm is true, or both paradigms are true (under a more nuanced understanding). In retrospect, it is more constructive to develop an understanding as to which paradigms may be right under what conditions, as opposed to an all-or-nothing approach of arguing about a singular right paradigm. Alternatively, we could come up with a new paradigm, a new school of thought, a new direction; which replaces or bridges the old way of thinking.

In this paper, we re-examine the current direction and paradigms of interpretability and invite the reader to consider whether it is time for a new paradigm.

\subsection{Interpretability and faithfulness}

Interpretability is the ability to explain a model in understandable terms to humans \citep{Doshi-Velez2017a}. Model explanations have particularly become important for AI safety, as machine learning is increasingly being used by the industry and affects the lives of most humans. This and additional motivations are elaborated on in \Cref{sec:why-interpretability}.

Within interpretability, there currently exist two paradigms, called \emph{post-hoc} and \emph{intrinsic} \citep{Lipton2018}. \Cref{sec:current-paradigms} properly describe their stance. Put briefly, the \emph{intrinsic} paradigm believes that only models designed to be explained can be explained \citep{Rudin2019};. In contrast, the \emph{post-hoc} paradigms believe this constraint is unnecessary and too restrictive to achieve competitive performance \citep{Madsen2021}.

The position in this paper is that neither paradigm has been fruitful because their underlying beliefs are problematic and we should therefore look for new directions. \Cref{sec:why-new-paradigm} contains the primary support for this position. To prove that other paradigms can exist, \Cref{sec:new-paradigms} then presents three emerging paradigms for interpretability and discusses how they might overcome past challenges, their beliefs, drawbacks, and future directions. Importantly, while this demonstrates that developing new paradigms is possible, \Cref{sec:new-paradigms} should not be considered a final list of directions.

At the core of this discussion is how each paradigm approaches faithfulness. A faithful explanation means the explanation accurately reflects the model's logic, and ensuring and validating this often presents a major challenge because the model's logic is inaccessible to humans \citep{Jacovi2020}. Faithfulness is particularly important, as false but convincing explanations can lead to unsupported confidence in models, increasing the risk of AI. 

New paradigms, such as those described in \Cref{sec:why-new-paradigm}, bring new perspectives regarding how to achieve faithfulness. This creates a new opportunity to do interpretability research centered around ensuring faithfulness. However, it also creates a new risk as we may take faithfulness for granted once again, as has been the case with both the intrinsic \citep{Jacovi2020} and post-hoc paradigms \citep{Madsen2021}. To prevent this, this paper also takes the position that we should be vigilant about faithfulness when it comes to these new paradigms to prevent repeating past mistakes.

\section{Why interpretability is needed}
\label{sec:why-interpretability}

Before discussing the current paradigms and their shortcomings, it's necessary to first consider if interpretability is needed at all. Many ethical motivations for interpretability are also served by bias and fairness metrics, so if the current paradigms of interpretability do not work (as we argue in \Cref{sec:why-new-paradigm}), perhaps we should drop the idea of interpretability completely. If the models can be made accurate, unbiased, and fair enough, do we need to explain the models? In this section, we will argue that interpretability is required, by examining the limitations of bias and fairness metrics and the scientific motivations for interpretability.

\subsection{Limitations of bias and fairness metrics}

There is no doubt that bias and fairness metrics present a vital role in validating models' behavior. However, a shared limitation is that they always measure known attributes \citep{Barocas2019}. For example, gender-bias metrics use gender attributes. This presents two challenges. Can we procure such attributes (known as protected attributes)? How do we prevent unanticipated biases? 

\subsubsection{Protected attribute procurement}
Attributes like gender, race, age, disability, etc., are under U.S. law known as ``protected attributes'' \citep{Xiang2019}, and collecting and using these attributes is heavily regulated in most of the world. \citet{Andrus2021} writes, ``In many situations, however, information about demographics can be extremely difficult for practitioners to even procure.''. Therefore, systematically measuring bias and fairness is not always practical \citep{Andrus2021}.

On the other hand, explanations often don't depend on knowing these protected attributes in advance and can provide a more qualitative analysis. For example, suppose an explanation tells us that the word ``Woman'' from ``Member of Woman's Chess Club'' in a resume is important for making a hiring recommendation. In that case, there is a potential harmful bias \citep{Kodiyan2019}. Therefore, explanations can serve a similar practical purpose to a fairness or bias metric without performing systematical correlations.

\subsubsection{Unknown attribute bias}
Although protected attributes are important to consider and are often legally protected, many more relevant attributes are involved in ensuring a fair and unbiased system. Unfortunately, it is impossible to consider every possible bias in advance. As an alternative, interpretability offers a more qualitative and explorative validation.

Continuing the example with resumes and automated hiring recommendations, during investigations by \citet{Fuller2021}, the authors found that a hospital only accepted candidates with computer programming experience when they needed workers to enter patient data into a computer. Another example was a clerk position where applicants were rejected if they did not mention floor-buffing (i.e., a cleaning method for floors) \citep{Fuller2021a}.

These examples present cases of systematic unintended bias. However, they do not relate to any protected attributes, and they are so specific they can only be discovered through qualitative explanations and investigations. That said, systematic fairness/bias metrics can quantify the damage once potential biases are identified using interpretability. Afterwards, those metrics can be integrated into a quality assessment system to prevent future harm.

\subsection{Interpretability for scientific discovery and understanding}

Interpretability is not only used for ethics and adjacent purposes, where bias and fairness metrics have an important role. Interpretability is also used for scientific discovery and learning about what makes models work.

\subsubsection{Scientific Discovery}
An example of scientific discovery is interpretability in drug discovery \citep{Preuer2019,Jimenez-Luna2020,Dara2022}. A common approach is to use importance measures to identify regions in genomic sequences responsible for a particular behavior, such as producing a protein. While these explanations do not guarantee that such connections exist in reality, they can provide important initial hypotheses for scientists enabling them to make more informed choices about the direction of their research.

\subsubsection{Model understanding}
An emerging field of interpretability is mechanistic interpretability, which identifies parts of a neural network that have a particular responsibility \citep{Cammarata2020}. For example, identifying a collection of neurons responsible for copying content in a generative language model, etc. \citep{Elhage2021}. Such insights may not be directly relevant to downstream tasks, but they help us understand current model limitations and can lead to better model design.

\section{The current paradigms of interpretability}
\label{sec:current-paradigms}

\begin{table*}[tb!]
    \centering
    \begin{tikzpicture}[
    every text node part/.style={align=center},
    rowsep/.style={color=black!40},
    title/.style={rectangle, fill=blue!8, anchor=center},
    content/.style={rectangle, fill=black!2, anchor=center},
]
\pgfmathsetmacro{\tcolgap}{0.7}
\pgfmathsetmacro{\tboxwidth}{7.4}
\pgfmathsetmacro{\trowtitlewidth}{0.65}
\pgfmathsetmacro{\colttitleheight}{0.70}
\pgfmathsetmacro{\tdefintionheight}{1.65}
\pgfmathsetmacro{\tbeliefheight}{1.5}
\pgfmathsetmacro{\ttextwidth}{21em}

\newcommand{\rAs}{\rBe} \newcommand{\rAe}{\rAs + \tdefintionheight}
\newcommand{\rBs}{\rCe} \newcommand{\rBe}{\rBs + \tbeliefheight}
\newcommand{\rCs}{0} \newcommand{\rCe}{\rCs + \tbeliefheight}

\newcommand{\cAs}{\tcolgap} \newcommand{\cAe}{\cAs + \tboxwidth}
\newcommand{\cBs}{\cAe + \tcolgap} \newcommand{\cBe}{\cBs + \tboxwidth}

\newcommand{\ttitle}[2]{
    \fill[title] (\csname c#1s\endcsname, \rAe) rectangle node[text width=\ttextwidth]{\textbf{#2}} (\csname c#1e\endcsname, \rAe + \colttitleheight);
}
\newcommand{\tcontent}[3]{
    \fill[content] (\csname c#2s\endcsname, \csname r#1s\endcsname) rectangle node[text width=\ttextwidth]{#3} (\csname c#2e\endcsname, \csname r#1e\endcsname);
}

\ttitle{A}{Intrinsic paradigm}
\tcontent{A}{A}{The model is designed to provide explanations by making the explanation part of the model architecture.}
\tcontent{B}{A}{Only models that were designed to be explained can be explained.}
\tcontent{C}{A}{Intrinsic models can have the same performance as a black-box model.}

\ttitle{B}{Post-hoc paradigm}
\tcontent{A}{B}{The model is produced without regard for explanation, and the explanations are then created after model training.}
\tcontent{B}{B}{Although it may be very challenging, black-box models can be explained.}
\tcontent{C}{B}{Black-box models will be more generally applicable than intrinsic models.}

\draw[rowsep] (\cAs - \trowtitlewidth, \rAe) -- (\cBe, \rAe);
\path (\cAs, \rAe) -- (\cAs, \rAs) node[midway, above, rotate=90]{\vphantom{$a^{a}_b$}defintion};
\draw[rowsep] (\cAs - \trowtitlewidth, \rBe) -- (\cBe, \rBe);
\path (\cAs, \rBe) -- (\cAs, \rCs) node[midway, above, rotate=90]{\vphantom{$a^{a}_b$}underlying beliefs};
\draw[rowsep] (\cAs - \trowtitlewidth, \rCs) -- (\cBe, \rCs);

\path (\cBe, \rAe) -- (\cBe + \trowtitlewidth, \rAe);

\end{tikzpicture}
    \vspace{1em}
    \caption{Comparison of the definitions and underlying beliefs of the intrinsic and post-hoc paradigms. The beliefs relate to a) requirements for a faithful explanation and b) model capabilities. It should be apparent that these two views are seemingly incompatible.}
    \label{fig:old-paradigms}
\end{table*}

This paper uses a common definition of interpretability, ``the ability to explain or to present in understandable terms to a human'' by \citep{Doshi-Velez2017a}. However, even this definition of interpretability is not agreed upon.

Lipton says, ``the term interpretability holds no agreed upon meaning, and yet machine learning conferences frequently publish papers which wield the term in a quasi-mathematical way'' \citep{Lipton2018}. In 2017, a UK Government House of Lords review of AI noted after substantial expert evidence that ``the terminology used by our witnesses varied widely. Many used the term transparency, while others used interpretability or explainability, sometimes interchangeably'' \citep[91]{HouseofLords2017}.

For this reason, there are also no clearly agreed-upon definitions of the current paradigms of interpretability \citep{Carvalho2019, Flora2022}. As such, this section defines the \emph{intrinsic} and \emph{post-hoc} paradigms, as well as describe their underlying beliefs, which are summarized in \Cref{fig:old-paradigms}.

\subsection{Definitions}
\citet{Jacovi2020} write: ``A distinction is often made between two methods of interpretability: (1) interpreting existing models via post-hoc techniques; and (2) designing inherently interpretable models. \citep{Rudin2019}''. Based on this and other sources \citep{Madsen2021,Arya2019,Carvalho2019,Murdoch2019}, this paper refers to these two ideas as 1) the \emph{intrinsic} paradigm and 2) the \emph{post-hoc} paradigm.

\subsubsection{The intrinsic paradigm}
The intrinsic paradigm works on creating so-called \emph{inherently interpretable models}. These models are architecturally constrained, such that the explanation emerges from the architecture itself.

Classical examples are decision trees or linear regression. In the field of neural networks some examples are: 1) ``Old-school'' attention \citep{Bahdanau2015,Jain2019}, where attention points to which input tokens are important. 2) Neural Modular Networks \citep{Andreas2016, Gupta2020, Fashandi2023}, which produce a prediction via a sequence of sub-models, each with known behavior. 3) Prototypical Networks \citep{Bien2009,Kim2014,Chen2019a}, which predicts by finding similar training observations.

\begin{figure}[H]
    \centering
    \includegraphics[width=\figurewidth,page=2]{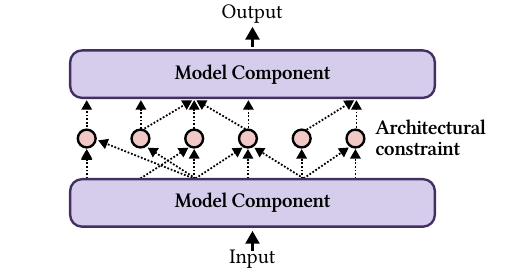}
    \caption{Abstract diagram of the intrinsic paradigm, where the model is architecturally constrained, such that the constraint itself is the explanation. In cases of Decision Trees the entire model is constrained, but often (e.g. Prototype Networks or Attention) only part of the model is constrained.}
    \label{fig:paradigm:intrinsic}
\end{figure}

\subsubsection{The post-hoc paradigm}
\emph{Post-hoc} explanations are computed after the model has been trained. They are developed independently of the model's architecture and how it was trained. However, there are often some simple criteria, like ``the model should be differentiable'', ``the training dataset is known'', or ``inputs are represented as tokens'' \citep{Madsen2021}. Although general applicability is technically not a requirement, if a method is so specific that it only works on one specific model, it's likely an \emph{intrinsic explanation}.

\begin{figure}[h]
    \centering
    \includegraphics[width=\figurewidth,page=1]{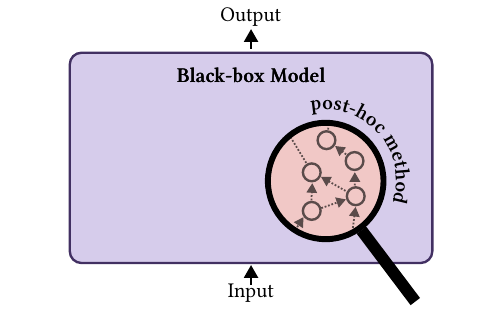}
    \caption{Abstract diagram of the post-hoc paradigm, where a post-hoc method is used to explain a black-box model. The post-hoc method is usually an algorithm, like the gradient w.r.t. the input, but it can also be an auxiliary model.}
    \label{fig:paradigm:post-hoc}
\end{figure}

As an example, a common post-hoc explanation is gradient-based importance measures. Importance measures explain which input features (words, pixels, etc.) are important for making a prediction. This is archived by differentiating the prediction with respect to the input. The idea is that if a small change in input causes a big change in the output, then that input is important \citep{Baehrens2010,Seo2018,Karpathy2015}.

\subsection{Beliefs}
As with all paradigms, there are fundamental underlying beliefs, which are why the paradigm's followers partake in their paradigm of choice. At the core of these beliefs are two central questions. When are explanations faithful and what are the requirements for faithfulness? And, how do these requirements affect the model's general performance capabilities?

\subsubsection{When are explanations faithful?}
The intrinsic paradigm believes that: \emph{only models designed to be explained, can be explained}, which their \emph{inherently interpretable models} try to satisfy. Therefore, they argue that using black-box models is too risky, as these models can never be faithfully explained \citep{Rudin2019}.

However, although their models are designed to be intrinsically explainable, this claim and their faithfulness should still be questioned \citep{Jacovi2020}, as many inherently interpretable model ideas are later revealed not to provide faithful explanations. For example, attention-based explanations have received notable criticism for not being faithful \citep{Jain2019,Serrano2019,Vashishth2019,Meister2021a,Madsen2022}. This is discussed more in \Cref{sec:why-new-paradigm:intrinsic}.

The \emph{post-hoc explanation} paradigm takes a less strict stance and believes that even models that were not designed to be explained (i.e., black-box models) can still be explained. However, as this paradigm has no control over the model, achieving faithful explanations is very challenging; this is discussed more in \Cref{sec:why-new-paradigm:post-hoc}.

In conclusion, the intrinsic paradigm considers explanations to be part of the model design, and post-hoc explanations are always applied after the model design. \citet{Madsen2021} frame intrinsic as proactive and post-hoc as retroactive. Hence, the two schools of thought are incompatible frameworks, and they can philosophically be considered as paradigms \citep{Kuhn1996}.

\subsubsection{What is the effect on the model's general performance capabilities?}

It would seem that \emph{intrinsic explanation} is the obvious choice. If we can control the model such that the faithfulness of explanations can be guaranteed, why consider \emph{post-hoc explanation}?

The commonly mentioned idea is that the \emph{post-hoc} paradigm believes that by constraining the models in the manners that the \emph{intrinsic paradigm} requires, there is a trade-off in performance \citep{DARPA2016}. However, this trade-off does not have to be the case in practice \citep[section 2]{Rudin2019}.

A more accurate take, which is rarely explicitly discussed, is that the common industry prefers off-the-shelf general-purpose models and only later thinks about interpretability \citep{Bhatt2020}. Additionally, most research only considers predictive performance, not interpretability. Therefore, \emph{intrinsic} researchers are always catching up to black-box models. From the \emph{post-hoc} perspective, it would make more sense to work on generally applicable interpretability methods for both off-the-shelf and future black-box models.

From the intrinsic perspective, while the industry might prefer off-the-shelf models now, they shouldn't. Not validating models through intrinsic explanations can have serious consequences \citep{Rudin2019} and eventually damage their business. Additionally, with increasing legal requirements to provide explanations, the industry may have to use inherently explainable models \citep{Goodman2017}.

For these reasons, the \emph{intrinsic} paradigm believes we should not let the industry's needs dictate our research direction, as their goals may be too short-sighted. In the long run, intrinsic models may be the only reasonable option.

In conclusion, the \emph{post-hoc} paradigm has good intentions with providing general explanations for general-purpose models. However, from the \emph{intrinsic} paradigm perspective, those good intentions are meaningless if it is fundamentally impossible to provide guaranteed faithful explanations without an \emph{inherently interpretable model}.

\section{Why interpretability needs a new paradigm}
\label{sec:why-new-paradigm}

It tends to be the case that when there are multiple paradigms, it is because neither of the paradigms fits the needs. However, for the case of the \emph{post-hoc} and \emph{intrinsic} paradigms, it could be argued that they serve different needs. For example, \emph{intrinsic} explanations should be preferred for critical applications \citep{Rudin2019}, and \emph{post-hoc} explanations could be used for verifiable situations, such as drug discovery, where the hypothesis generated by the explanations is verified using physical experiments.

\subsection{The case against the intrinsic paradigm}
\label{sec:why-new-paradigm:intrinsic}
The industry primarily uses post-hoc explanations, including for high-stakes applications such as insurance risk assessment and financial loan assessment \citep{Bhatt2020,Krishna2022}. This is because such industries usually do not have the in-house expertise to develop custom inherently interpretable models. They must rely on basic inherently interpretable models, like decision trees, which are not competitive, or use more advanced off-the-shelf neural black-box models, like pre-trained language models, which will be competitive. In practice, the industry is thus not in a position to choose inherently interpretable models.

Another challenge with the intrinsic paradigm is that its models are often not completely interpretable because only a part of the model is architecturally constrained to be interpretable. The rest of the neural network, still use black-box components (e.g. Dense layer, Recurrent layer, etc.) which are not interpretable. As such, the intrinsic promise should not be taken at face value \citep{Jacovi2020}.

An example of this is classic attention-based models \citep{Bahdanau2015, Jain2019}. Attention itself is interpretable, as it's a weighted sum, and explains the importance of each intermediate representation. However, attention is often used as token-importance. This does not work, as the intermediate representations are produced by a black-box recurrent neural network (e.g. LSTM \citealt{Hochreiter1997}) which can mix or move the relationship between tokens and the intermediate representations. Therefore, the attention scores do not necessarily represent token-importance \citep{Bastings2020}.

Likewise, Neural Modular Networks produce an executable problem composed of sub-networks, such as \texttt{find-max-num(filter(find()))}, which is interpretable.\citep{Fashandi2023,Andreas2016,Gupta2020}. However, each sub-networks (\texttt{find-max-num}, \texttt{filter}, \texttt{find}) is itself a black-box model with little guarantee that it operates as intended \citep{Amer2019, Subramanian2020, Lyu2024}.

Overall, there are few success stories with intrinsic explanations. They are either not performance-wise competitive, general-purpose enough for the industry \citep{Bhatt2020}, or their intrinsic claims are unsupported \citep{Jacovi2020}.

\subsection{The case against the post-hoc paradigm}
\label{sec:why-new-paradigm:post-hoc}
Although post-hoc explanations directly address the interpretability challenge of black-box components and models, and could therefore provide more complete explanations, there are also few success stories with post-hoc, where post-hoc explanations are consistently faithful.

Most notable is perhaps post-hoc importance measure (IM) explanations, where the explanation indicates which input features are the most important for making a prediction. The pursuit of such explanations has produced countless papers \citep{Binder2016,Ribeiro2016,Li2016a,Shrikumar2017,Smilkov2017,Sundararajan2017a,Ahern2019,Thorne2019,ElShawi2019,Sangroya2020}.

However, repeatedly, the faithfulness of these IM explanations is criticized \citep{Adebayo2018, Adebayo2022, Kindermans2019,Hooker2019,Slack2020,Yeh2019}. For example, there is great disagreement between alleged faithful IMs, which is hard to reconcile \citep{Jain2019,Krishna2022}. Other works show their faithfulness is both task and model-dependent and thus don't provide the generality that the \emph{post-hoc} paradigm desires \citep{Bastings2021, Madsen2022}. Finally, theoretical works suggest that IMs are subject to a \emph{no free lunch theorem} \citep{Han2022}, or it may be impossible to provide faithful post-hoc IMs \citep{Bilodeau2024}.

Similar to the work of IM, is visualization of neurons in computer vision, which shows that neurons represent high-level concepts, such as nose or dog. This is done by visualizing convolutional weights or the input image that maximizes a neuron's activation \citep{Olah2017,Nguyen2016,Yosinski2015}, which provides very convincing evidence. However, it has been shown empirically, theoretically, and through human-computer-interaction (HCI) studies that these visualizations do not provide useful explanations regarding the neurons' responsibility \citep{Geirhos2023,Borowski2021,Zimmermann2021}\footnote{Neural networks likely do encode high-level concepts, but these visualizations are not useful for identifying the responsibility of specific neurons.}.

Another notable example is probing explanations, where models are verified by relating the model's behavior or intermediate representation to, for example, linguistic properties (part-of-speech, etc.) \citep{Belinkov2019, Belinkov2020}. This idea has produced an entire subfield called BERTology \citep{Rogers2020}. BERTology in particular has attained substantial attention \citep{Coenen2019,Clark2019a,Rogers2020,Clouatre2021a,ThomasMcCoy2020,Conneau2018,Tenney2019a}, with most of the works finding that neural networks can learn linguistic properties indirectly.

Unfortunately, like post-hoc importance measures, there are many reasons to be highly skeptical \citep{Belinkov2021}. For example, using an untrained model or a randomized dataset shows an equally high correlation with linguistic properties, compared with training a regular model \citep{Zhang2018, Hewitt2019}. These discoveries have put the entire methodology into question, although there is work trying to adapt to these new critiques \citep{Voita2020}.

\subsection{Summary}
Post-hoc importance measures and probing explanations are just two cases where post-hoc shows initial promise through countless papers, only to be debunked repeatedly. The trend is oscillating between proposing new explanation methods and debunking them. Of course, it's impossible to prove that there will never be a great post-hoc method. However, the lack of guarantees also makes it impossible to know when a faithful post-hoc method is proposed. Similarly, intrinsic explanations also receive criticism after a while, as has been the case with attention and Neural Modular Networks.

\begin{table*}[t!]
    \centering
    \begin{tikzpicture}[
    every text node part/.style={align=center},
    rowsep/.style={color=black!40},
    title/.style={rectangle, fill=blue!8, anchor=center},
    content/.style={rectangle, fill=black!2, anchor=center},
]
\pgfmathsetmacro{\tcolgap}{0.4}
\pgfmathsetmacro{\tboxwidth}{5.1}
\pgfmathsetmacro{\trowtitlewidth}{0.6}
\pgfmathsetmacro{\colttitleheight}{1.05}
\pgfmathsetmacro{\tdefintionheight}{1.65}
\pgfmathsetmacro{\tbeliefheight}{1.65}
\pgfmathsetmacro{\ttextwidth}{14em}

\newcommand{\rAs}{\rBe} \newcommand{\rAe}{\rAs + \tdefintionheight}
\newcommand{\rBs}{\rCe} \newcommand{\rBe}{\rBs + \tbeliefheight}
\newcommand{\rCs}{0} \newcommand{\rCe}{\rCs + \tbeliefheight}

\newcommand{\cAs}{\tcolgap} \newcommand{\cAe}{\cAs + \tboxwidth}
\newcommand{\cBs}{\cAe + \tcolgap} \newcommand{\cBe}{\cBs + \tboxwidth}
\newcommand{\cCs}{\cBe + \tcolgap} \newcommand{\cCe}{\cCs + \tboxwidth}

\newcommand{\ttitle}[2]{
    \fill[title] (\csname c#1s\endcsname, \rAe) rectangle node[text width=\ttextwidth]{\textbf{#2}} (\csname c#1e\endcsname, \rAe + \colttitleheight);
}
\newcommand{\tcontent}[3]{
    \fill[content] (\csname c#2s\endcsname, \csname r#1s\endcsname) rectangle node[text width=\ttextwidth]{#3} (\csname c#2e\endcsname, \csname r#1e\endcsname);
}

\ttitle{A}{Learn-to-faithfully-explain paradigm}
\tcontent{A}{A}{The model is optimized such that an explanation method becomes faithful.}
\tcontent{B}{A}{The relaxed faithfulness metric used for optimization is a sufficient approximation.}
\tcontent{C}{A}{Models can be optimized such that explanations become faithful without losing performance.}

\ttitle{B}{Faithfulness measurable model paradigm}
\tcontent{A}{B}{The model is designed to enable measuring faithfulness of a category of explanations.}
\tcontent{B}{B}{It is computationally feasible to optimize explanations for optimal faithfulness.}
\tcontent{C}{B}{Models can be optimized to be faithfulness measurable without loss of predictive performance.}

\ttitle{C}{Self-explaining model paradigm}
\tcontent{A}{C}{The model directly outputs both its prediction and an explanation for that prediction.}
\tcontent{B}{C}{Models can be trained to model and articulate their own reasoning accurately.}
\tcontent{C}{C}{Self-explanation capabilities do not negatively impact regular predictions.}

\draw[rowsep] (\cAs - \trowtitlewidth, \rAe) -- (\cCe, \rAe);
\path (\cAs, \rAe) -- (\cAs, \rAs) node[midway, above, rotate=90]{\vphantom{$a^{a}_b$}defintion};
\draw[rowsep] (\cAs - \trowtitlewidth, \rBe) -- (\cCe, \rBe);
\path (\cAs, \rBe) -- (\cAs, \rCs) node[midway, above, rotate=90]{\vphantom{$a^{a}_b$}underlying beliefs};
\draw[rowsep] (\cAs - \trowtitlewidth, \rCs) -- (\cCe, \rCs);

\path (\cCe, \rAe) -- (\cCe + \trowtitlewidth, \rAe);

\end{tikzpicture}
    \vspace{1em}
    \caption{Comparison of the definitions and underlying beliefs of the new paradigms. The beliefs relate to a) explanation requirements and b) model capabilities. These new paradigms can be compared with the old paradigms in \Cref{fig:old-paradigms}.}
    \label{fig:new-paradigms}
\end{table*}

\section{Are new paradigms possible?}
\label{sec:new-paradigms}
Although both the intrinsic and post-hoc paradigms have significant issues, parts of their underlying beliefs have merit. The intrinsic paradigm believes that \emph{we can't expect models that were not designed to be explained, to be explained}, while post-hoc believes \emph{black-box models tend to be more general purpose while providing high predictive performance}. These beliefs have merit, and it's worth considering how to incorporate their spirit into new paradigms.

It can seem unlikely that such a paradigm can even exist. However, there is already some work that satisfies these desirables. In particular, we have identified 3 alternative paradigms, summarized in \Cref{fig:new-paradigms}. These directions are all fairly new and unfortunately have not received much focus, likely due to favoritism towards existing paradigms \citep{Kuhn1996}.

All three mentioned paradigms work with what-would-be black-box models. However, their idea is to optimize these models, such that they are designed to be explained. How they differ, is in their exact formulation of this approach.

It's important to note, that it's only with hindsight we can truly know if a new idea will become the next major paradigm, and it may be a fourth unknown idea that will become the next major paradigm. As such, the main purpose of this section is not to promote the next paradigm but rather to establish that it is possible to develop new interpretability paradigms.

\subsection{The learn-to-faithfully explain paradigm}
This paradigm is the most direct application of the optimization idea. An existing explanation algorithm \citep{Bhalla2023} or model is used \citep{Yoon2019,Chen2018}, and the predictive model is then optimized to maximize both the predictive performance and the faithfulness.

Importantly, this approach does not require the architectural constraints that the intrinsic paradigm applies, as the explanation comes from an external explanation method, not the architectural design. The explanation method can be similar or even identical to those from the post-hoc paradigm. However, because the model is optimized to enable these explanations to be faithful, it's not post-hoc, and there are more reasons to think that the explanations should be faithful.

\begin{figure}[h]
    \centering
    \includegraphics[width=\figurewidth,page=3]{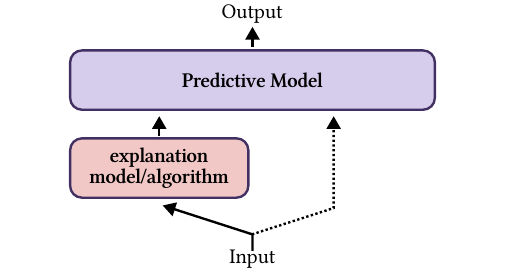}
    \caption{Abstract diagram of the learn-to-faithfully explain paradigm. In most cases, this paradigm works by generating an explanation from the input, using either a model or an algorithm, this explanation is then fed into the predictive model, which has been optimized to respect the explanation.}
    \label{fig:paradigm:learn-to-explain}
\end{figure}

Early work on this jointly trains an explanation model and a prediction model \citep{Yoon2019,Chen2018}. This direction has been called joint amortized explanation methods (JAMs). However, \citet{Jethani2021} point out that the explanation model often learns to encode the prediction, which means the explanation model becomes part of the black-box problem rather than the solution. A solution can be to use a disjoint setup \citep{Jethani2021}, where the explanation model can't encode the prediction, a setup that following works have adapted \citep{Jethani2022,Covert2023}. However, the explanation model may still output unfaithful explanations for out-of-distribution inputs. An alternative is to produce the explanation algorithmically \citep{Bhalla2023}, for example by having an explanation algorithm remove unnecessary features, and the prediction model learns to support sparse features.

Regardless of the specific approach used to produce the explanation, the challenges are formalizing the faithfulness objective correctly such that the optimization works as intended, ensuring that the explanations are truly faithful and that the model properties that make explanations faithful also hold for out-of-distribution data \citep{Covert2023, Bhalla2023}.


\subsection{The faithfulness measurable model paradigm} 
This paradigm integrates measuring the faithfulness of an explanation into the model design, such faithfulness can be easily measured without requiring architectural constraints. This can be of a huge advantage, as measuring faithfulness is often extremely challenging \citep{Jacovi2020}. Importantly, because faithfulness is easy to measure by design, it's possible to identify the explanation that maximizes faithfulness using optimization algorithms \citep{Zhou2022a}, which makes the model indirectly intrinsically explainable \citep{Madsen2023,Hase2021,Vafa2021}. In essence, this paradigm reformulates the intrinsic paradigm from `inherently explainable'' to ``inherently measurable''. 

\begin{figure}[h]
    \centering
    \includegraphics[width=\figurewidth,page=4]{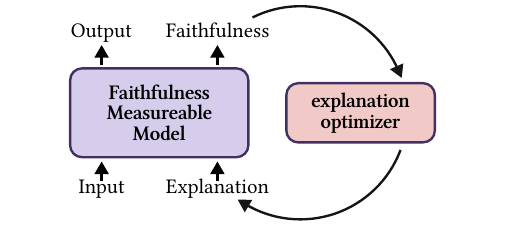}
    \caption{Abstract diagram of the faithfulness measurable model paradigm. In this paradigm, the predictive model can also measure how faithful a given explanation is. The explanation can thus be produced by optimizing an initial (maybe random) explanation towards maximal faithfulness.}
    \label{fig:paradigm:fmm}
\end{figure}

\citet{Madsen2023} and \citet{Hase2021} show that this idea can be achieved using simple data argumentation, and there is no need for architectural constraints. The central idea is to use the erasure metric \citep{Samek2017} to measure faithfulness of importance measures. The erasure metric says: if information (pixels, tokens, etc.) is truly important, then when removing it the prediction should change significantly. The common challenge is that removing information causes out-of-distribution issues \citep{Hooker2019,Madsen2022}. However, by using data argumentation during training, it's possible to extend the model to support the partial inputs created by the erasure metric. Importantly this can be achieved without architectural constraints, thus it remains possible to use general-purpose models such as RoBERTa \citep{Madsen2023} and GPT-2 \citep{Vafa2021}.

The challenge in this paradigm is about coming up with a way to integrate the faithfulness metric in the model, while ensuring there is no performance impact and that the model operates in-distribution \citep{Madsen2023}. Additionally, developing efficient optimization procedures for optimizing explanations is difficult, due to the discrete nature of many explanations \citep{Hase2021,Zhou2022a}.


\subsection{The self-explaining model paradigm}
Rather than using external algorithms or models to produce explanations, \citet{Elton2020} proposes in this paradigm that models should explain themselves, meaning they become \emph{self-explaining}. The most common implementation of this idea is instruction-tuned large language models (e.g., ChatGPT, Gemini, etc.)  \citep{OpenAI2023,Jiang2023,Touvron2023}, which are allegedly able to explain themselves in great detail and very convincingly \citep{Chen2023,Agarwal2024}.

\begin{figure}[h]
    \centering
    \includegraphics[width=\figurewidth,page=5]{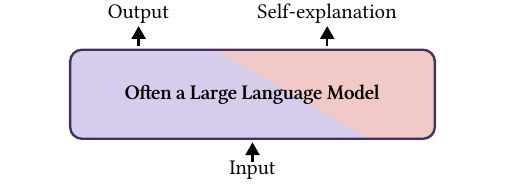}
    \caption{Abstract diagram of the self-explanation paradigm, where the same model is trained to produce both the regular predictive output and an explanation, called a self-explanation. This paradigm is often seen with Large Language Models, where both the predictive output and the self-explanations appear as generated text.}
    \label{fig:paradigm:self-explain}
\end{figure}

Because the explanations are produced by a black-box this paradigm can be quite dangerous. Therefore, there must be solid evidence that the explanations are faithful for this approach to be valid. However, despite this immediate danger, the model that generates the explanation can in principle have access to all of the logic that produces the prediction. At a minimum, the same weights produce both the prediction and the explanation.

Importantly, self-explanations must relate to the model's reasoning logic, not just the world or abstract concepts. However, presently there is little evidence that this is satisfied \citep{Turpin2023a,Lanham2023,Madsen2024}. This is not surprising, as the self-explanations are explicitly trained based on humans' annotating how these explanations should look. However, humans don't have any insight into how the model operates \citep{Jacovi2020}. As such, the model converges towards very convincing self-explanations with no regard for faithfulness \citep{Agarwal2024,Chen2023}.

While the other paradigms have found some solutions to their challenges, there are currently no known solutions to make self-explanations faithful. Even measuring faithfulness of self-explanation is very challenging \citep{Huang2023}. Currently there only exist a few metrics for specific self-explanations \citep{Parcalabescu2023}. However, future work may improve upon this, by developing more faithfulness metrics and by aligning models not just towards human preference but also faithfulness and truthfulness.

\section{Limitations}
As with any other position paper, this paper presents one position: that we must seek out new paradigms of interpretability. However, one could make other valid positions or directions about the future of interpretability. 

In particular, this position paper primarily focuses on faithfulness. However, as interpretability is about how we can explain models to humans in understandable terms, one must also consider how well-understandable an explanation is. This concern is known as human-groundedness \citep{Doshi-Velez2017a}, simulatability \citep{Lipton2018}, or comprehensibility \citep{Robnik-Sikonja2018a}.

This concern is orthogonal to the faithfulness concern and is therefore not discussed in this paper. However, readers are encouraged to study recent works like \citet{Schut2023,Kim2022}, which propose the new idea that it is not enough to frame explanations in terms that humans already understand. We should also develop new language and mental abstractions for humans to understand machines.

\section{Conclusion}

Although some evidence exists for the new paradigms presented in \Cref{sec:new-paradigms}, these are, first and foremost, just ideas. It's only in retrospect that we can truly know if one paradigm results in meaningful progress in the field. It is also entirely possible that neither of these ideas is what moves the interpretability field forward.

For these reasons, the core position of this paper is that we should develop new directions and paradigms in interpretability, instead of focusing on the existing post-hoc and intinsic paradigms, which are currently dominating.

That being said, we must also be vigilant regarding faithfulness to avoid repeating past mistakes \citep{Jacovi2020}. These new paradigms will present new arguments for why their method is faithful. As we are unfamiliar with these arguments, identifying their flaws is difficult, and it will be easy to get swayed by them.

Historically, a common tactic in post-hoc works was convincing visualization of explanations that aligned with our intuitions \citep{Olah2017,Yosinski2015,Nguyen2016}. However, such visualizations are empty arguments, as humans can't know what a true explanation looks like \citep{Geirhos2023}. Likewise, intrinsic works have made seemingly strong theoretical arguments for why their methods are faithful, but these arguments failed to capture the whole model. Even the new \emph{learn-to-faithfully-explain paradigm} have already shown sharp corners, where the explainer model unintentionally encode the prediction and is therefore unfaithful \citep{Jethani2021}.

To prevent false arguments, a sound start is to always have a specific and measurable definition of faithfulness, which works for all methods within a given explanation category (e.g., counterfactual or importance measures).

Finally, while these new paradigms are promising, it's unlikely they will completely erase the current paradigms. We still teach both the particle and wave paradigms in physics. Most scientists don't worry about whether there are true statements in math that cannot be proven.

Likewise, there will likely always be situations where intrinsic or post-hoc interpretability makes sense. For example, basic statistics and linear regressions can be framed as intrinsic interpretability. Hence, if a company or researchers decide to use a model because of its intrinsically explainable properties, then we should only praise them -- as long as they also measure the faithfulness of the explanations.

\section*{Acknowledgements}
Sarath Chandar is supported by the Canada CIFAR AI Chairs program, the Canada Research Chair in Lifelong Machine Learning, and the NSERC Discovery Grant.

Siva Reddy is supported by the Facebook CIFAR AI Chairs program and NSERC Discovery Grant.

\bibliography{references}
\bibliographystyle{latex/icml2024}

\end{document}